\documentclass[aps,prx,preprint,tightenlines,notitlepage,floatfix,amsmath,amssymb]{revtex4-2}
\usepackage[T1]{fontenc}
\usepackage{microtype,mathtools,etoolbox}
\usepackage{graphicx}
\usepackage[dvipsnames]{xcolor}
\usepackage{needspace}
\usepackage{enumitem}
\usepackage{xurl}
\definecolor{ink}{HTML}{183344}
\definecolor{teal}{HTML}{007F7B}
\definecolor{rust}{HTML}{BF6035}
\usepackage[colorlinks=true,linkcolor=ink,citecolor=teal,urlcolor=teal]{hyperref}
\hypersetup{pdftitle={Counterexample-driven autonomous research with application to excess radiated energy in a Z-pinch},pdfauthor={Bowen Zhu and Jian Wu}}
\setlist{nosep,leftmargin=1.5em}
\makeatletter
\preprintsty@sw{\newcommand{\colfigwidth}{3.1in}\patchcmd{\frontmatter@abstract@produce}{\do@output@MVL}{\addvspace{\frontmatter@preabstractspace}\@gobble}{}{}}{%
 \newcommand{\colfigwidth}{\linewidth}%
 \renewcommand{\section}{\@startsection{section}{1}{\z@}%
  {16pt plus 2pt minus 2pt}{6pt}%
  {\normalfont\small\bfseries\centering}}
 \renewcommand{\subsection}{\@startsection{subsection}{2}{\z@}%
  {12pt plus 2pt minus 2pt}{5pt}%
  {\normalfont\small\bfseries\centering}}
 \renewcommand{\subsubsection}{\@startsection{subsubsection}{3}{\z@}%
  {10pt plus 2pt minus 2pt}{4pt}%
  {\normalfont\small\itshape\centering}}
 \setlength{\parskip}{0pt}
}
\makeatother
\graphicspath{{figures/}}
\newcommand{\simj}{\textsc{Simjecture}}
\newcommand{\dd}{\mathrm{d}}
\newcommand{\code}[1]{\texttt{#1}}

\begin{document}
\title{Counterexample-driven autonomous research with application to excess radiated energy in a Z-pinch}
\author{Bowen Zhu}
\author{Jian Wu}
\email{jxjawj@mail.xjtu.edu.cn}
\affiliation{State Key Laboratory of Electrical Insulation and Power Equipment, Xi'an Jiaotong University, Xi'an 710049, China}
\date{\today}

\begin{abstract}
Recent breakthroughs in artificial-intelligence-assisted mathematics show that open-ended machine search can outperform human mathematicians, yet transferring this paradigm to physics requires refactoring the research workflow. We introduce \simj{}, an open-source framework in which language-model agents pursue human-defined physical conjectures. The framework tackles two central challenges in this transfer: first, because physics is empirical and no model is final, it structures inquiry around counterexample-driven falsification and prospective hypothesis repair; second, because physical theories must interact with the real world, it uses computational physics as an executable testbed while separating the qualification of numerical instruments from scientific claim review so solvers and diagnostics can be adapted within a recorded investigation. We apply the framework in a $136.3$~h investigation to tackle the long-standing problem of radiated energy exceeding the imploding-shell kinetic energy in a Z-pinch. The investigation reveals competing effects of energy conversion, retention, and release timing. Stronger coherent helical deformation increases resistive conversion while reducing radiation and increasing retained energy. In a matched fine-grid test, broadening the trailing-plasma profile delays the later radiation maximum by $1.80$~ns and recovers an early $153$~J bolometric deficit to within $0.08\%$ by $44$~ns, with a spectral response that shifts during the pulse. These results provide new insight into energy conversion in high-energy-density plasma physics and demonstrate a new paradigm for physics research.
\end{abstract}
\maketitle

\section{Introduction}
\label{sec:introduction}

\subsection{From mathematical discovery to computational physics}

Recent breakthroughs in artificial intelligence (AI)-assisted mathematics have demonstrated the ability of AI to fill out and even penetrate beyond the convex hull of human knowledge. Over a span of months in 2026, automated search helped construct planar point configurations with at least $n^{1+\delta}$ unit-distance pairs ($\delta>0$) that disproved Erd\H{o}s's longstanding $n^{1+o(1)}$ unit-distance conjecture~\citep{alon2026unitdistance}, uncovered a three-dimensional polynomial counterexample to the Jacobian conjecture whose constant nonzero Jacobian determinant and non-injectivity can be checked directly~\citep{anthropic2026jacobian,gao2026jacobian}, and extended to mathematical physics through a reported finite-time velocity blowup for the three-dimensional forced incompressible Navier--Stokes equations accompanied by Lean formalizations and broader proof catalogues~\citep{openai2026navierstokes,openai2026nslean,openai2026mathcatalogue}. These developments share a common characteristic: AI excels at tasks with a strong asymmetry between discovery and verification, where finding a decisive construction requires expansive search while checking a concrete candidate gives an independent verifier a well-defined task.

Translating this paradigm to physics introduces two additional obligations. First, physics is empirical and no model is final: it is extremely difficult to prove a theory universally right, whereas finding a counterexample is much simpler. Second, physical theories must ultimately answer to nature through observation and experiment; however, laboratory campaigns are costly and rarely repeatable under identical microscopic conditions. Counterexample-driven computational physics investigations therefore provide an executable model world in which the driving motive is simple and controlled interventions can be repeated and dissected.

Yet every computational physics investigation rests on two intertwined foundations: a \emph{mathematical and numerical core} that integrates the chosen equations numerically, and a \emph{physical core} that selects the governing equations, closures, and boundary conditions used to represent the real system. When a simulation appears to violate a physical conjecture, that discrepancy can originate either from genuine model physics or from numerical effects. Only after the mathematical and numerical core is shown to be reliable can a contradiction falsify a physical claim, motivate a revised hypothesis with new testable predictions, and guide laboratory experiments. \simj{} reflects this duality directly by coupling two distinct workflows: \emph{scientific investigation}, which tests and revises physical conjectures, and \emph{instrument development}, which builds, modifies, and qualifies the numerical solvers and diagnostics required for those tests. The framework delegates solver engineering, numerical qualification, and controlled physical comparisons so agents can revise solvers and diagnostics as an investigation develops.

\subsection{Autonomous research with humans in the loop}

Much of the recent wave of autonomous research in machine learning instead treats inquiry as end-to-end manuscript generation. Karpathy's \emph{autoresearch} exemplifies the metric-driven loop across repeated language-model training runs~\citep{karpathy2026}, while \emph{The AI Scientist} automates idea generation, coding, plotting, and paper writing~\citep{lu2026aiscientist}. These systems arrive as conference and preprint ecosystems are already strained: NeurIPS main-track submissions more than tripled from 9,467 in 2020 to 30,709 in 2026~\citep{neurips2025review,neurips2026stats}, and arXiv introduced per-author monthly submission caps and stricter moderation rules in 2026 after monthly submissions reached 40,363 amid a surge of low-quality automated preprints~\citep{arxiv2025report,arxiv2026policy}.

End-to-end manuscript automation carries three fundamental scientific risks. First, an agent can game a convenient proxy without touching the underlying physical mechanism. Second, research ideas selected autonomously by AI agents rarely hold up as good scientific taste once executed~\citep{si2025ideation}. Third, automated pipelines tend to gravitate toward narrow, closed-loop studies that leave little room for future improvement or lasting scientific value.

As demonstrated in mathematics and recent physics-focused agentic systems for optics, scattering, accelerator control, and model exploration~\citep{krenn2021theseus,landgraf2025autoscatter,neudecker2025experiments,naegele2026sciexplorer}, automated workflows have proved most illuminating when anchored to well-posed domain questions and conjectures. We therefore argue that \emph{scientific taste}, namely judging which physical conjecture merits sustained compute and what evidence would genuinely change our physical picture, remains a human responsibility. Accordingly, an autonomous research system requires an open-box design so that its decisions remain understandable to human scientists, and its scientific reports must be re-examined by human researchers to resolve disputed or inconclusive results. Human steering should be an integral part of the scientific research system: the goal of an autonomous research system should not be to remove humans from the loop, but to design for effective human--AI collaboration.

\section{Research framework}
\label{sec:framework}

\subsection{Conjectures, decision rules, and minimal repair}

To investigate a physical conjecture efficiently, the system first translates the natural-language question into an explicit mathematical statement. A conjecture $H=(M,\mathcal D,Q,P)$ fixes a physical model $M$ specifying the governing equations and closures, a domain $\mathcal D$ of admissible simulation setups $x$ together with the physical regime in which the claim is meant to hold, a target observable $Q$, and a predicted range $P(x)$ for that observable, whether testing an internal prediction of $M$ or agreement with experimental data. Solved exactly with model $M$, the ideal conjecture asserts
\begin{equation}
 Q(x)\in P(x)\qquad\text{for every } x\in\mathcal D.
\end{equation}

A simulation, however, never returns the exact value $Q(x)$. It measures $Q$ through a numerical instrument $I$ comprising the solver, grid, and diagnostic scripts, and after mesh-refinement, timestep, and diagnostic controls the instrument returns a measurement range $I_Q(x)$. Along the \emph{instrument route}, the numerical setup $I$ is commissioned and refined through convergence, conservation, and calibration tests tailored to $Q$. If a capability is missing or a numerical control fails so that $I_Q(x)$ cannot be trusted, the instrument must be revised and requalified ($I\to I'$) before any scientific claim is evaluated; the physical conjecture $H$ remains open. Once the instrument is qualified, a setup $x_\star\in\mathcal D$ is an accepted counterexample only if this entire range lies outside the prediction:
\begin{equation}
 I_Q(x_\star)\cap P(x_\star)=\varnothing.
 \label{eq:qualifiedcounterexample}
\end{equation}
Comparing $I_Q(x)$ against $P(x)$ across the tested setups yields one of three outcomes:
\begin{itemize}
 \item \emph{Falsified:} a setup $x_\star\in\mathcal D$ satisfies Eq.~\eqref{eq:qualifiedcounterexample}, establishing a counterexample that disproves $H$.
 \item \emph{Supported within the tested scope:} every tested setup $x\in\mathcal D$ satisfies $I_Q(x)\subseteq P(x)$, supporting $H$ across the sampled regime.
 \item \emph{Unresolved:} either $I_Q(x_\star)$ straddles the boundary of $P(x_\star)$, or a numerically verified observable combines multiple physical processes without isolating the mechanism named in $H$. When the evidence leaves $H$ unresolved, the investigation may seek a diagnostic that isolates the quantity named in $H$, or use the findings to formulate a related hypothesis with a distinct observable and a prediction specified before its test; testing that hypothesis advances the investigation while leaving the status of $H$ unchanged.
\end{itemize}

Along the \emph{scientific route}, when $H$ is falsified, a revised hypothesis $H'=(M',\mathcal D',Q,P')$ can be formulated. Simply widening $P'(x)$ to cover observed extrema or deleting the failed regime from $\mathcal D'$ is trivially true without adding physical insight; \emph{minimal repair} therefore requires an explicit physical justification for every changed assumption, scope restriction, or bound in $(M,\mathcal D,P)$ while preserving unaffected predictions and accounting for the parent counterexamples. Before testing $H'$, a prospective commitment fixes the revised prediction $P'(x)$, domain $\mathcal D'$, and planned test cases: only fresh runs executed after this commitment can validate $H'$, while the parent conjecture $H$ and its counterexamples remain permanently recorded in the hypothesis tree even if candidate repairs subsequently fail.

\subsection{Research roles and scientific tools}
\label{sec:knowledge}

\begin{figure}[tbp]
 \centering
 \includegraphics[width=\colfigwidth]{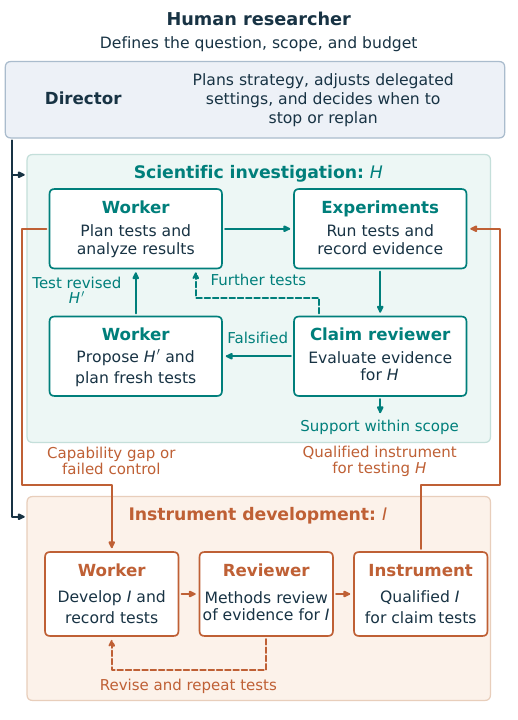}
 \caption{Coupled scientific (teal) and instrument (orange) workflows in \simj{}. }
 \label{fig:architecture}
\end{figure}

\simj{} implements these two review routes through the roles shown in Fig.~\ref{fig:architecture}. The human researcher defines the question and scope. A \emph{director} plans campaign strategy, monitors scientific coverage and compute budget, adjusts operational limits and helper-worker allocation, and decides when to stop stalled runs, replan, or call on the human scientist for steering, while an autonomous \emph{worker} configures simulations, inspects fields, writes code, and searches for counterexamples. Every worker submission along the instrument (orange) and scientific (teal) routes is gated by independent \emph{reviewers} operating in isolated model contexts. The methods reviewer checks diagnostic definitions, conservation accounting, and mesh or timestep controls to verify that the instrument measures the target observable. The claim reviewer then examines whether the conclusion follows within the declared scope and whether proposed repairs survive prospective tests.

Guided by tool-specific skills, the worker can select and combine installed tools as the physical question requires, including using one solver's output as input to another. Public demonstrations~\citep{simjecture} illustrate both a \href{https://drawingsword.com/simjecture/demos/qin1-hardware-plasma.html}{Qin-1 hardware-design study}, in which a magnetic field computed with Elmer is transferred to WarpX~\citep{vay2018} as a static background for a hybrid particle-in-cell calculation, and a \href{https://drawingsword.com/simjecture/demos/flash-warpx-patch.html}{fluid-to-kinetic study}, in which a FLASH current-sheet snapshot initializes a local WarpX calculation. Because computational campaigns on a given facility also build cumulatively across studies, related investigations share a persistent workspace that indexes project instructions, solver documentation, input configurations, prior conversations, and completed study reports. Both human researchers and autonomous agents query this library via retrieval-augmented generation~\citep{lewis2020}; revision-tracked citations and frozen launch contexts keep cross-study reuse subject to the current study's evidence contract.

\subsection{Open-box execution, minimal evidence harness, and research supervision}
\label{sec:openbox}
\label{app:implementation}

Beyond combining existing tools, resolving a computational physics question often requires modifying the instrument itself. \simj{} uses an open-box design in which the worker can inspect raw multi-dimensional fields, write new diagnostics, and modify initializers or physics modules as the physical question evolves. The system automatically records every code modification, input configuration, and simulation output so that the reviewer can audit whether a change alters only numerical settings, tests a new setup within $\mathcal D$, redefines a diagnostic, or modifies the governing model $M$ itself.

To keep open-box execution auditable while leaving the worker free to design experiments and use native coding tools, a minimal host harness enforces three service calls: \code{lab.run()} executes a calculation, \code{lab.commit()} registers a prospective repair plan, and \code{lab.review()} requests an independent verdict. At study launch, the host freezes the baseline evidence contract, permitted instruments, resource budget, and completion policy: \emph{answer} closes once the root conjecture or a qualified repair is supported or falsified, whereas \emph{repair} requires a validated replacement whenever the root conjecture is falsified. For each local or remote job, \code{lab.run()} snapshots the submitted source tree, input declarations, command bindings, and runtime identities before execution and records the output cryptographic hashes after the job finishes. Reference solver installations and human-supplied commissioning files remain read-only while the worker develops and qualifies code changes on an isolated branch. Because the worker has no interface path to approve its own claims, \code{lab.review()} verifies artifact hashes, hypothesis lineage, and post-commitment timestamps before passing the evidence packet to an isolated reviewer. During multi-day campaigns, the director can replan or seek human steering at stage boundaries, while a small supervisory policy, Conjecture~1~\citep{conjecture1}, can route routine status checks without deciding the scientific verdict.

\section{Validation study: reconnection-rate scaling}
\label{sec:cases}
\label{sec:island}

We first apply the framework to a controlled question in two-dimensional compressible resistive magnetohydrodynamics (MHD) island coalescence in FLASH~\citep{fryxell2000}. The human researcher supplies the physical hypothesis, a working FLASH application, and initial diagnostics, while the worker selects the test points and qualifies the measurements. The conjecture predicts that the local log-slope of the reconnection rate remains within $[-0.55,-0.45]$ around the Sweet--Parker $S_\eta^{-1/2}$ reference for resolved, pre-plasmoid sheets across the nominal Lundquist-number range $S_\eta=L_0 v_{A0}/\eta=1/\eta\in[250,4000]$, where $\eta$ is the magnetic diffusivity and the reference length and Alfv\'en speed satisfy $L_0=v_{A0}=1$ in normalized code units. The rate is measured across a fixed interval of reconnected magnetic flux, with exact definitions and numerical controls detailed in Appendix~\ref{app:islandmethods}.

An exploratory $128^2$ scan indicates that the reconnection rate falls more slowly with $S_\eta$ than the reference scaling, and the prospectively declared largest-departure rule selects the contrast $S_\eta=1000\to2000$ for targeted qualification. Because a stable integrated flux rate can coexist with an under-resolved local current sheet as diffusivity decreases, the worker audits current-sheet profiles and magnetic topology at matched reconnected flux alongside mesh, timestep, sampling-cadence, and independent flux-reconstruction controls. When an intermediate refinement step ($256^2\to384^2$ at $S_\eta=2000$) increases the rate shift, methods review blocks claim evaluation and requests a finer $512^2$ calculation; only after the instrument passes review does the worker execute fresh production trajectories.

\begin{figure}[tbp]
 \centering
 \includegraphics[width=\colfigwidth]{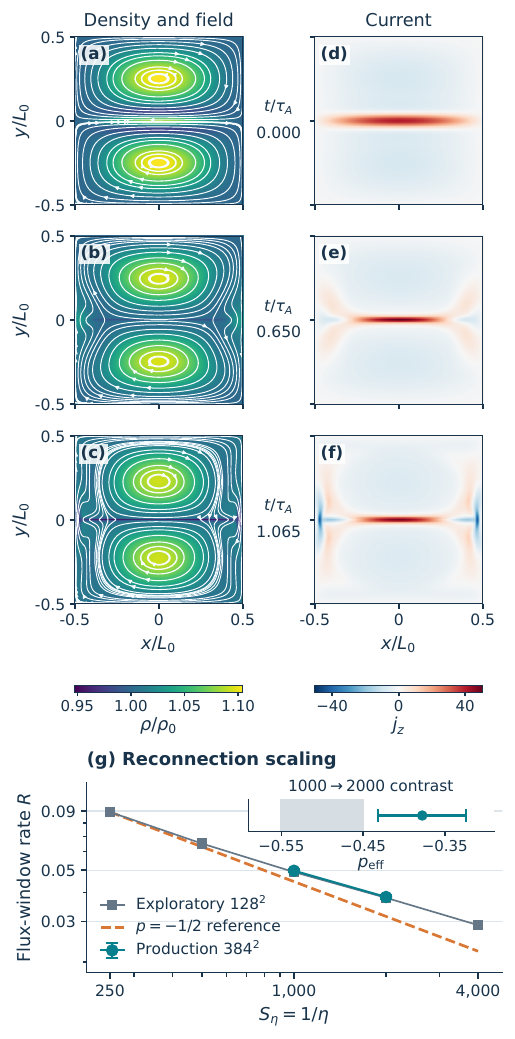}
 \caption{FLASH island-coalescence study. (a)--(c) Density and magnetic streamlines and (d)--(f) out-of-plane current $j_z$ at $S_\eta=1000$ ($384^2$ grid) for $t/\tau_A=0$, $0.6502$, and $1.0651$ ($\tau_A=L_0/v_{A0}$; $L_0$, $B_0$, and $v_{A0}$ are reference length, magnetic field, and Alfv\'en speed). (g) Exploratory ($128^2$) and qualified production ($384^2$) reconnection rates versus $S_\eta$, with Sweet--Parker reference $R_{\rm ref}\propto S_\eta^{-1/2}$ (dashed) and inset comparing the qualified slope interval $[-0.4314,-0.3247]$ against the predicted band $[-0.55,-0.45]$.}
 \label{fig:islandfields}
\end{figure}

The fresh $384^2$ production runs yield a local scaling exponent $p_{\rm eff}=-0.3778$ with an empirical numerical sensitivity interval $[-0.4314,-0.3247]$ (Fig.~\ref{fig:islandfields}(g)), which lies entirely above $-0.45$ and falsifies the proposed scaling band [Eq.~\eqref{eq:qualifiedcounterexample}]. Throughout the measurement window, the magnetic-null inventory retains a single X point and two O points in every saved field snapshot, confirming the pre-plasmoid regime at the resolved scales, and independent claim review reproduces the rates and error propagation from the recorded outputs to accept the counterexample.

After falsifying the $[-0.55,-0.45]$ band, the worker proposes two candidate repairs: a transient sheet-width model driven by the measured inflow strain and upstream magnetic field, and an amplitude-budget extension. Without an adjustable rate prefactor, both models underpredict the anchor rates by $25\%$--$31\%$ and fail their prospective calibration checks. The workflow therefore records both failed repair attempts and closes the study with a verified counterexample and an open theoretical question; a companion FLASH-to-WarpX fluid-to-kinetic current-sheet study in the archived repository~\citep{simjecture} illustrates the complementary outcome in which a bounded hypothesis is supported across numerical and seed controls.

\makeatletter\preprintsty@sw{}{\Needspace{14\baselineskip}}\makeatother
\section{Investigation of radiated-energy excess in a Z-pinch}
\label{sec:alradiation}

We now apply \simj{} to a central open problem in pulsed-power plasma physics: determining how much of the radiated X-ray energy excess above the implosion kinetic energy in a stagnating wire-array Z-pinch is supplied by magnetic reconnection. Because the reconnection share is unknown a priori, we pose an initial working conjecture that non-axisymmetric magnetic reconnection directly accounts for at most $10\%$ of the positive radiation excess above a reference implosion kinetic energy, using this quantitative threshold to launch a four-stage, $136.3$~h human--AI collaborative investigation in three-dimensional radiation-MHD. On the fine mesh, total Ohmic heating reaches $12.74\%$ of the radiation excess. However, the volume-integrated $\eta_e J^2$ diagnostic combines reconnecting and non-reconnecting dissipation, leaving the $10\%$ conjecture unresolved. Testing this conjecture nonetheless reveals how magnetic conversion, energy retention, and release timing compete. Stronger coherent helical deformation boosts three-dimensional resistive conversion but weakens radial compression and traps more magnetic and thermal energy. Conversely, broadening the trailing-plasma profile shunts current outward and delays the stagnation radiation peak before late emission recovers the early bolometric deficit.

\subsection{Physical background and model}
\label{sec:almodel}

Fast wire-array Z-pinches convert pulsed electromagnetic energy into imploding plasma motion, thermal energy, and intense soft and K-shell X-ray bursts~\citep{ryutov2000,haines2011}. Across both multi-megampere facilities and $1$~MA university-scale drivers, the radiated X-ray energy frequently exceeds the maximum instantaneous bulk kinetic energy of the imploding load~\citep{cuneo2005,ivanov2013}, reflecting that instantaneous kinetic energy captures only a transient stage of energy transfer in a continuously driven system. As the core stagnates, magneto-Rayleigh--Taylor ($m=0$) and kink ($m=1$) instabilities redistribute current between the dense column and surrounding trailing plasma~\citep{felber1981,chittenden2004}, while Lorentz-force work, compression, shocks, rapid cross-field ion heating~\citep{stollberg2023}, and resistive dissipation contribute simultaneously to the energy budget~\citep{pontin2022}.

To anchor the model in a well-diagnosed regime, we follow an eight-wire aluminum array on the $1$~MA Zebra pulsed-power generator: $15~\mu\mathrm{m}$ wire diameter, $20$~mm axial length, and $76.34~\mu\mathrm{g}$ total mass~\citep{ivanov2013}. In this $1$~MA regime, laser shadowgraphy, Faraday rotation, X-ray diodes, and three-dimensional simulations directly resolve the coexistence of a dense stagnating core, trailing plasma, and current redistribution, while aluminum ($Z=13$) provides well-characterized equations of state and opacities alongside a distinct $>1$~keV hard-band emission channel above sub-keV thermal radiation. To study late implosion and radiative stagnation without modeling early cold-wire ablation, the calculation begins from a prescribed pre-ablated coaxial distribution: an inward-moving Gaussian core shell centered at $0.8$~mm with a $100~\mathrm{km/s}$ speed parameter weighted by the local core fraction, surrounded by a nearly stationary trailing annulus centered at $2.5$~mm with equal nominal mass fractions. This post-ablation parameterization isolates stagnation-phase conversion and current shunting from early wire-ablation uncertainties while allowing the column geometry and trailing-plasma breadth to be varied independently at fixed mass and drive.

FLASH~4.8~\citep{fryxell2000} evolves three-temperature ($T_i,T_e,T_{\rm rad}$) resistive MHD coupled to tabulated aluminum equations of state and opacities, electron--ion exchange, thermal conduction, implicit magnetic diffusion, and 30-group flux-limited radiation diffusion. Three-dimensional Cartesian calculations and axisymmetric $r$--$z$ references use a constant $1$~MA drive, periodic axial boundaries, and transverse circuit/open boundaries through which multigroup radiation escapes, with numerical schemes, domain geometry, and baseline profile parameters detailed in Appendix~\ref{app:alnumerics}.

\subsection{Campaign progression overview}
\label{sec:alscreening}

Because the investigation spans dozens of baseline, screening, nonlinear, and diagnostic-validation runs across two geometries and three initializer generations, we introduce a compact naming convention to track individual calculations across the timeline and tables. Each label encodes the domain geometry, transverse resolution, and seed family: \code{3D-N} gives the cell count $N$ along each transverse Cartesian axis, whereas \code{RZ-N} gives the radial cell count in axisymmetric cylindrical geometry. For three-dimensional runs, \code{Lp} denotes an initial density-modulation amplitude of $p\%$, while \code{S} and \code{M} denote the \emph{Single-helix} and \emph{Multimode} initializers, where the suffix $d/w$ specifies a centerline displacement $D=0.1d$~mm, defined by the root-mean-square (RMS) amplitude $D_{\rm rms}$ for \code{M}, and a trailing-plasma radial width $\sigma_t=0.1w$~mm. Appending \code{-P} marks a short pilot or control run used to qualify an initializer, verify a diagnostic, or test execution settings. A parenthesized Roman numeral identifies the campaign stage for repeated physical setups. Full mesh dimensions and nonlinear parameters are tabulated in Appendix~\ref{app:alnumerics}.

\begin{figure*}[tbp]
\centering
\includegraphics[width=\textwidth]{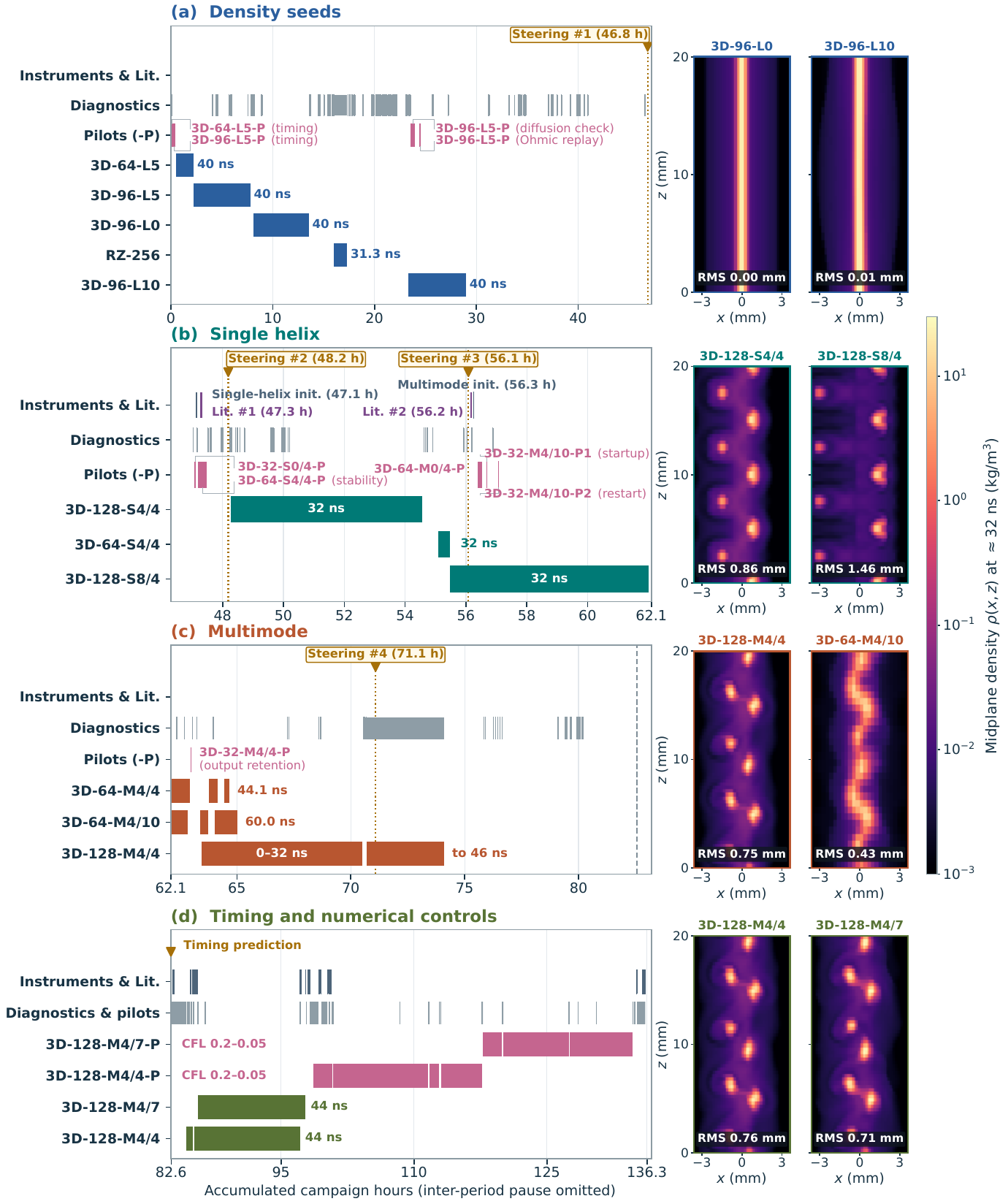}
\caption{Investigation progression through (a) density-seed screening, (b) single-helix tests, (c) multimode comparisons, and (d) timing and numerical controls. Bars show recorded job intervals in accumulated campaign hours; horizontal scales differ between panels. Right-hand panels show $x$--$z$ density slices near $32$~ns on a shared scale. RMS denotes the mass-weighted displacement of axial slice centroids from the original axis.}
\label{fig:alscreening}
\end{figure*}

Prior to autonomous execution, the workspace was seeded with two unperturbed commissioning baselines (\code{RZ-64} and \code{3D-32-L0}), a partial \code{RZ-256} checkpoint, a Message Passing Interface (MPI) domain-decomposition benchmark, and baseline diagnostic scripts. The investigation progressed through four experimental stages over $136.3$~h in two execution periods (Fig.~\ref{fig:alscreening}(a)--(d)). The timeline accumulates their recorded durations, omitting the pause between Stages~III and~IV.

\textbf{Stage~I: Density-seed screening.} In Stage~I (Fig.~\ref{fig:alscreening}(a)), the worker first ran short timing pilots (\code{3D-64-L5-P}, \code{3D-96-L5-P}) to gauge computational cost and then executed four principal three-dimensional screening trajectories across mesh size and seed amplitude alongside the axisymmetric continuation \code{RZ-256}. Despite refining the mesh and raising the density perturbation to $10\%$, the pinch column remained virtually straight through stagnation (compare \code{3D-96-L0} and \code{3D-96-L10} in Fig.~\ref{fig:alscreening}(a)), yielding negligible non-axisymmetric Ohmic heating and almost unchanged bolometric radiation. Two \code{3D-96-L5-P} checkpoint-replay pilots served as diagnostic checks that compared Ohmic heating reconstructed from the electron resistivity $\eta_e$ and current density $J=|\mathbf J|$ as $\eta_e J^2$ against the native magnetic-diffusion ledger and prompted an extended cycle of harmonic and current-sheet diagnostic scripts. At the end of this stage, because the pinch column remained overly stable with minimal three-dimensional deformation, Human Steering~\#1 directed the agent to consult the Z-pinch instability literature and construct a physically disruptive three-dimensional initialization.

\textbf{Stage~II: Single-helix tests.} Opening Stage~II (Fig.~\ref{fig:alscreening}(b)), the worker developed the \emph{Single-helix initializer} to displace the initial column centerline into a coherent $m=1$ helix coupled with $m=0$ axial necking, so that magnetic tension and hoop forces act directly on a bent current channel. In parallel, the first literature synthesis (\code{Lit.\,\#1}) motivated tracking both sausage and kink modes and separating initial geometric offset from dynamical amplification. Before launching production trajectories, the agent used two pilot jobs (\code{3D-32-S0/4-P} and \code{3D-64-S4/4-P}) for instrument and workflow qualification to verify baseline equivalence and short-window integration stability. Steering~\#2 then noted that three-dimensional unsplit corner-transport upwind integration runs stably at a Courant--Friedrichs--Lewy number of $\mathrm{CFL}=0.8$, accelerating integration fourfold. The resulting principal trajectories (\code{3D-128-S4/4}, the coarse resolution control \code{3D-64-S4/4}, and the doubled-amplitude run \code{3D-128-S8/4}) produced pronounced nonlinear helical kinking and beaded dense cores (Fig.~\ref{fig:alscreening}(b), right). Physically, comparing \code{3D-128-S8/4} against \code{3D-128-S4/4} revealed a striking trend: doubling the initial helical displacement reduced total radiated X-ray energy while leaving more magnetic and thermal energy stored in the plasma. While \code{3D-128-S8/4} was still executing, Steering~\#3 and a second literature review (\code{Lit.\,\#2}) noted that experimental wire-array implosions develop multi-wavelength axial structure embedded in ablated trailing plasma, prompting the worker to prepare the next solver generation in parallel.

\textbf{Stage~III: Multimode comparisons.} Built and qualified during the second half of Panel~(b) and executed in Stage~III (Fig.~\ref{fig:alscreening}(c)), the \emph{Multimode initializer} superposed three axial wavelengths with mixed helicities and exposed the trailing-plasma radial width $\sigma_t$ as an explicit control parameter. Before the main Stage~III runs, \code{3D-64-M0/4-P} verified baseline recovery, while \code{3D-32-M4/10-P1} and \code{P2} tested two-step initialization and checkpoint handoff. The worker then launched the concurrent coarse pair \code{3D-64-M4/4} and \code{3D-64-M4/10} to isolate the effect of broadening the trailing plasma; when a wrapper-level retention issue interrupted the end of their initial $32$~ns segment, the agent ran \code{3D-32-M4/4-P} to verify the output-retention fix before launching the fine multimode trajectory \code{3D-128-M4/4} alongside the two late-time coarse continuations. The resulting density fields (Fig.~\ref{fig:alscreening}(c), right) and radiation histories showed that both multimode seeding and trailing-plasma broadening reshape the timing and hard-X-ray ($>1$~keV) spectrum more strongly than the bolometric total. Steering~\#4 initiated a comprehensive audit of the completed Stage~I--III trajectories. The resulting physical synthesis in Secs.~\ref{sec:alpartition} and~\ref{sec:alresponse} motivated the next experimental stage.

\textbf{Stage~IV: Timing and numerical controls.} The current redistribution and late radiative recovery in Stage~III suggested testing an intermediate trailing width, $\sigma_t=0.7$~mm, against a matched $0.4$~mm reference on the fine mesh, with an expected $\sim2$--$3$~ns delay and subsequent radiative recovery (Fig.~\ref{fig:alscreening}(d)). This phase extended the physical inquiry to magnetic conversion, retention, and delayed release, using the completed trajectories and their interpretation as starting material. The worker evolved the fine pair through $44$~ns and found that absolute late-time thermal inventories remained sensitive to the timestep despite stable integration. It therefore branched checkpoint controls at progressively smaller timesteps and audited the radiation diagnostics. These tests established local contraction of the timestep response, while the retained physical trajectories allowed the intermediate-width timing prediction to be examined in detail.

\subsection{Radiation excess and energy conversion}
\label{sec:alpartition}

\begin{figure*}[tbp]
\centering
\includegraphics[width=\linewidth]{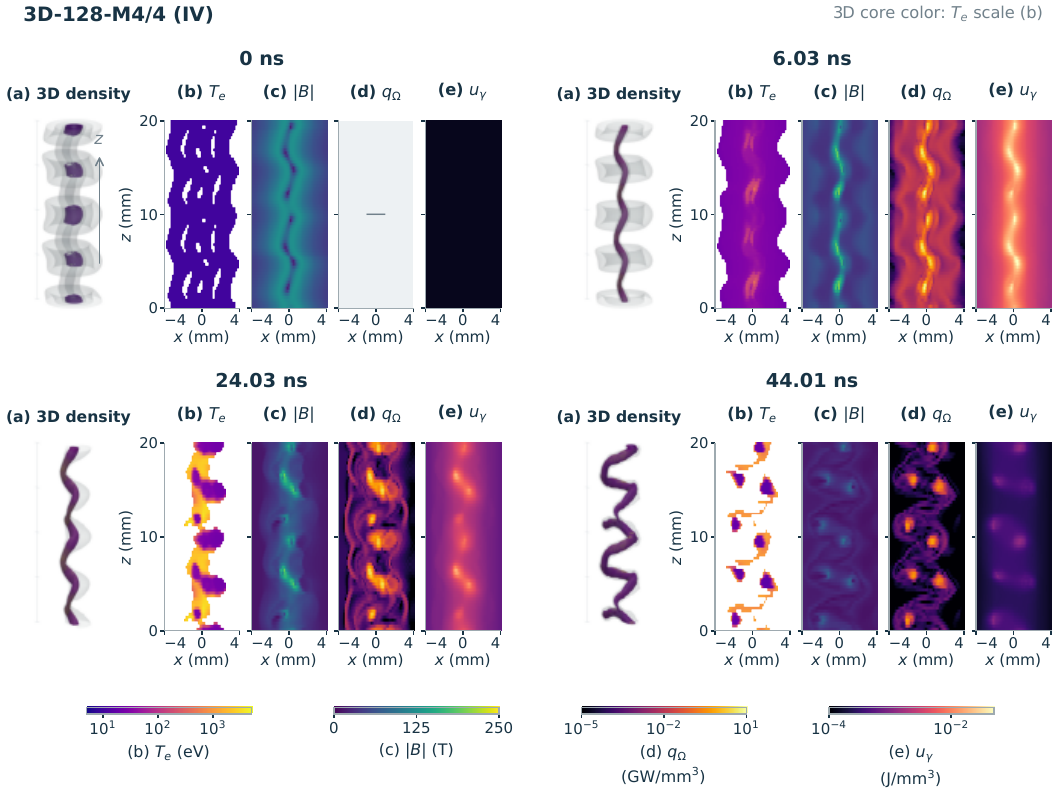}
\caption{Plasma evolution in the Stage~IV realization of \code{3D-128-M4/4}. Within each block, columns show (a) three-dimensional density surfaces at $0.1$ and $1~\mathrm{kg/m^3}$, with the inner surface colored by electron temperature using the scale in (b); (b) electron temperature; (c) magnetic-field magnitude; (d) Ohmic power density; and (e) stored radiation energy density. Field maps are longitudinal cuts at $y=0$, interpolated between the two bracketing cell-center planes. Temperature is shown where $\rho\ge0.01~\mathrm{kg/m^3}$.}
\label{fig:alstructure}
\end{figure*}

Figure~\ref{fig:alstructure} tracks the Stage~IV three-dimensional multimode implosion \code{3D-128-M4/4}. Magnetic hoop and tension forces gather the imploding shell into a helically bent filament. Its fastest-converging segments produce an initial radiation flash at $5.56$~ns as current shifts outward into the lower-density corona. Overall radial compression then continues until the centroid-relative radius reaches its $0$--$32$~ns minimum at $25.27$~ns, by which point $87.4\%$ of the $0$--$32$~ns radiation yield has already escaped. Near this compression minimum, the longitudinal $y=0$ cut intersects the continuously connected three-dimensional column as a chain of radiatively cooled dense patches near $T_e\sim 30$~eV. Immediately outside this cool core, azimuthal magnetic fields, Ohmic dissipation, and stored radiation concentrate in a lower-density peripheral sheath around a $0.4$--$4.0$~keV hot spot, before the column expands into larger helical excursions by $44.01$~ns.

Across this stagnation sequence, the radiation yield exceeds the shell kinetic reference by a large margin. Over the fixed $3.149$--$25.029$~ns evaluation window, \code{3D-128-M4/4} radiates $3.086$~kJ compared with a tagged-load kinetic energy of $K_{\rm ref}=0.498$~kJ evaluated at $9.776$~ns, the last retained field sample before the declared $10$~ns reference time. The radiation yield is $6.20$ times $K_{\rm ref}$, giving an excess of $E_{\rm excess}=2.588$~kJ.

The magnetic reservoir supplies most of the net decrease in stored energy over this interval. Separating the tagged aluminum load from the low-density numerical background, we track the load kinetic energy $K_{\rm L}$ and thermal energy $U_{g,\rm L}=U_{e,\rm L}+U_{i,\rm L}$, together with magnetic and radiation-field energies integrated over the whole domain. Stored magnetic energy $E_B$ falls by $2.498$~kJ, load kinetic energy falls by $0.389$~kJ, and load thermal energy falls by $24.6$~J. Together with stored radiation energy $U_\gamma$ and escaped radiation energy $E_{\gamma,\rm escape}$, these changes enter the balance
\begin{equation}
 E_{\gamma,\rm escape}=-\Delta E_B-\Delta K_{\rm L}-\Delta U_{g,\rm L}-\Delta U_\gamma+\mathcal B_{\rm L},
 \label{eq:albalance}
\end{equation}
where $\Delta$ denotes the change across the stated interval. Stored radiation $U_\gamma$ changes by less than $1$~J, and the inferred net exchange with the load and fields is $\mathcal B_{\rm L}=0.174$~kJ, as defined in Appendix~\ref{app:vacuumaccounting}.

To test the proposed reconnection contribution on this fixed interval $W$ across the domain volume $V$, we compare the reconstructed resistive conversion $Q_\Omega(W)$ with the radiation excess $E_{\rm excess}(W)$,
\begin{equation}
 \begin{aligned}
 Q_\Omega(W)&=\int_W\!\int_V\eta_eJ^2\,\dd V\,\dd t,\\
 E_{\rm excess}(W)&=E_{\gamma,\rm escape}(W)-K_{\rm ref}.
 \end{aligned}
 \label{eq:alenergy}
\end{equation}
On the fine mesh, the field reconstruction gives $Q_\Omega=0.330$~kJ, agreeing with the native diffusion increment and reaching $12.74\%$ of $E_{\rm excess}$.

\begin{figure}[tbp]
\centering
\includegraphics[width=\colfigwidth]{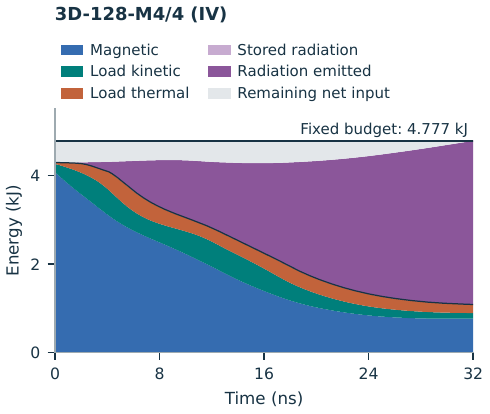}
\caption{Energy partition of the aluminum load and fields over $0$--$32$~ns in \code{3D-128-M4/4 (IV)}. Bands show magnetic energy, load kinetic and thermal energies, stored radiation, cumulative emitted radiation, and remaining inferred net input (gray). Thermal energy includes electrons and ions; magnetic and radiation-field energies cover the whole domain. The horizontal top is the fixed budget through $32$~ns; the lower dark curve traces stored load-and-field energy.}
\label{fig:alenergy}
\end{figure}

Across the broader $0$--$32$~ns common observation window, the local conversion pathways follow from the magnetic-energy equation,
\begin{equation}
 \frac{\partial}{\partial t}\frac{B^2}{2\mu_0}+\nabla\!\cdot\mathbf S=-\mathbf v\!\cdot(\mathbf J\!\times\mathbf B)-\eta_eJ^2,
 \label{eq:almagnetic}
\end{equation}
where $\mu_0$ is the vacuum permeability, $\mathbf S$ is the Poynting flux, and $\mathbf v$ is the plasma velocity. Magnetic forces drive motion that heats the plasma through compression and shocks, while resistive dissipation transfers magnetic energy directly into thermal energy. Figure~\ref{fig:alenergy} places magnetic, aluminum-load kinetic, aluminum-load thermal, and stored-radiation energies alongside cumulative radiation escape over $0$--$32$~ns in \code{3D-128-M4/4}. Combining the initial load-and-field energy with the $\mathcal B_{\rm L}(T)=0.485$~kJ net exchange inferred through $T=32$~ns defines a fixed budget of $4.777$~kJ, where the gray band $\mathcal B_{\rm L}(T)-\mathcal B_{\rm L}(t)$ tracks the remaining net input before $32$~ns. As the magnetic reservoir shrinks, the imploding load decelerates through stagnation and its kinetic energy drops to $0.124$~kJ at $32$~ns, leaving $1.084$~kJ stored in the load and fields, including $0.192$~kJ of load thermal energy, while $3.694$~kJ has escaped as radiation.

Comparing the matched single-helix runs \code{3D-128-S4/4} and \code{3D-128-S8/4} over $0$--$32$~ns further shows that higher resistive conversion does not increase the radiated yield. Doubling the initial helical displacement from $0.4$ to $0.8$~mm raises the $32$~ns RMS slice-centroid offset from $0.86$ to $1.46$~mm and increases the centroid-relative RMS radius $a_\perp$ from $0.686$ to $0.956$~mm. Stronger helical kinking thus weakens local radial compression and leaves an elongated, lower-density core. Consequently, although the native diffusion-stage gas-energy increment rises from $503$ to $622$~J, snapshot-integrated magnetic-force work decreases from approximately $3.35$ to $3.02$~kJ, escaping radiation falls by $197$~J, and final whole-domain stored energy rises by $182$~J, with combined whole-domain magnetic and thermal storage increasing by $209$~J.

These energy budgets establish three conclusions. First, the radiation excess draws on a magnetic energy reservoir much larger than the instantaneous shell kinetic energy, as magnetic forces and resistive dissipation continuously convert magnetic energy while photons escape. Second, although volume-integrated Ohmic heating reaches $12.74\%$ of the radiation excess on the fine mesh, this total cannot be attributed to magnetic reconnection alone because $\eta_e J^2$ includes non-reconnecting resistive diffusion and reconnection can also energize bulk flows~\citep{pontin2022}; isolating the reconnection share requires dedicated topology diagnostics and finer mesh convergence. Third, resistive heating, mechanical compression work, and magnetic and thermal storage evolve together, so increasing Ohmic dissipation while weakening transverse compression reduces the net radiated X-ray energy. Quantitative definitions of the geometric metrics, operator-level energy ledgers, resolution and window sensitivities, and work integrals across all cases are consolidated in Appendices~\ref{app:alnumerics} and~\ref{app:alenergychecks}.

\subsection{Instability and radiative response}
\label{sec:alresponse}

\begin{figure*}[tbp]
\centering
\includegraphics[width=\linewidth]{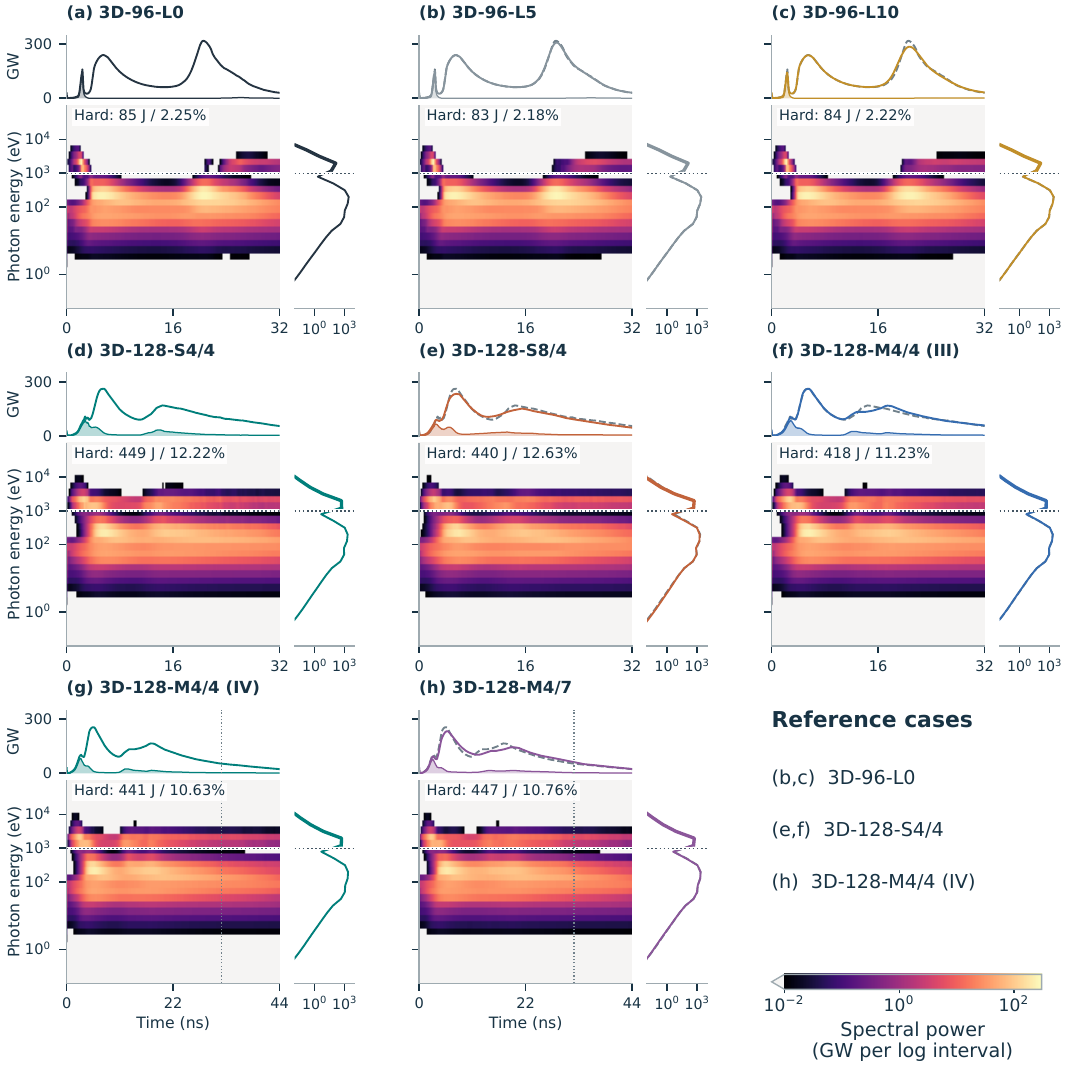}
\caption{Time-resolved radiation spectrograms, power histories, and integrated spectra for the unperturbed and density-seeded references (a)--(c), the fine-mesh deformation set from Stages~II--III (d)--(f), and the Stage~IV fine-mesh trailing-width pair (g),(h). Each set uses its own matched reference. Shaded power regions and bold spectral segments mark emission above $1$~keV; dashed curves show the indicated reference.}
\label{fig:alradiation}
\end{figure*}

Figure~\ref{fig:alradiation} compares how initial instabilities and radial mass profiles reshape the radiation pulse across the campaign. In the unperturbed column and the linear density-seed runs (Fig.~\ref{fig:alradiation}(a)--(c)), the $5\%$ and $10\%$ density modulations leave both the total radiated energy and the photon spectrum above $1$~keV nearly unchanged from the unperturbed reference. Across this baseline group, $78\%$ of the $>1$~keV emission escapes during the initial $0$--$8$~ns transient, followed by weak hard emission through $24$~ns.

\textbf{Centerline deformation.} Introducing explicit centerline deformations produces a stronger early pulse and sustains hard X-ray emission across stagnation (Fig.~\ref{fig:alradiation}(d)--(f)). Increasing the single-helix displacement from $0.4$ to $0.8$~mm weakens the first power peak, delays the main stagnation peak by about $1$~ns, and lowers the $0$--$32$~ns bolometric yield by $5.34\%$ while changing the $>1$~keV yield by only $2.17\%$. Replacing the single $0.4$~mm helix with the multimode perturbation breaks the uniform helical pitch into uneven bends and variable filament thickness, exciting strong higher azimuthal harmonics such as $m=3$ at late time. This geometric fragmentation leaves the bolometric yield within $1.07\%$ of the single-helix run yet reduces the $>1$~keV yield by $7.08\%$, with the matched $64^2\!\times32$ pair showing the same pattern of bolometric invariance within $0.18\%$ alongside a $4.74\%$ hard-band reduction. Single-mode displacement therefore lowers total radiated energy mainly by weakening bulk compression, whereas multimode fragmentation selectively softens the high-energy spectrum.

\begin{figure}[tbp]
\centering
\includegraphics[width=\colfigwidth]{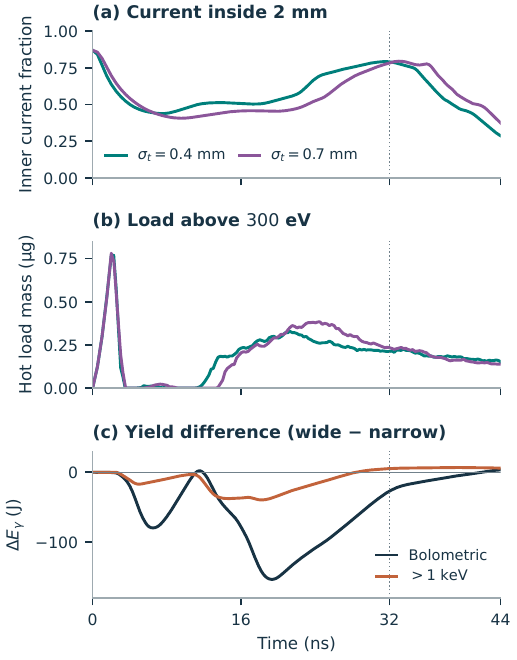}
\caption{Response of the matched fine-grid trailing-width pair, \code{3D-128-M4/4} and \code{3D-128-M4/7}. (a) Signed axial-current fraction within $2$~mm of the slice centroid. (b) Tagged load mass above $300$~eV. (c) Cumulative radiation difference, wider minus narrower profile, for all photons and for $>1$~keV photons. Dotted lines mark $32$~ns.}
\label{fig:alcurrentresponse}
\end{figure}

\textbf{Trailing-plasma width.} Broadening the trailing annulus redistributes axial current outward and delays radiation release, as shown for the fine-mesh pair $\sigma_t=0.4$ and $0.7$~mm (Figs.~\ref{fig:alradiation}(g),(h) and~\ref{fig:alcurrentresponse}). In the Stage~III coarse screen from $\sigma_t=0.4$ to $1.0$~mm, broadening the profile reduced the $12$--$24$~ns inner-current fraction from $0.536$ to $0.402$, delayed the stagnation power peak by $4.58$~ns, and recovered $60\%$ of the $32$~ns bolometric deficit by $44$~ns while lowering the cumulative $>1$~keV yield by $6.21\%$. Testing the intermediate width $\sigma_t=0.7$~mm on the fine mesh in Stage~IV reproduces the outward current redistribution and delayed radiation release: between $12$ and $24$~ns, the mean signed axial-current fraction within $2$~mm of the column centroid falls from $0.535$ to $0.460$ (Fig.~\ref{fig:alcurrentresponse}(a)), delaying the rise of the $>300$~eV tagged load by $1.64$~ns (Fig.~\ref{fig:alcurrentresponse}(b)). The measured stagnation-peak delay was $1.80$~ns, slightly below the anticipated $2$--$3$~ns range, and the predicted late radiative recovery was observed (Fig.~\ref{fig:alcurrentresponse}(c)). Specifically, the bolometric deficit reaches $153$~J near $19.3$~ns, narrows to $27$~J at $32$~ns, and ends $3.4$~J ($0.08\%$) above the $4.15$~kJ reference at $44$~ns after the wider profile emits $30.49$~J more over $32$--$44$~ns at $\mathrm{CFL}=0.4$ and $27.87$~J more at $\mathrm{CFL}=0.2$. Meanwhile, the $>1$~keV fraction shifts from $7.38\%$ versus $10.10\%$ over $8$--$16$~ns to $9.98\%$ versus $7.79\%$ over $16$--$24$~ns, leaving cumulative hard emission within $1.35\%$. This sequence of distributed current, delayed core heating, and late radiative catch-up parallels Zebra experiments in which trailing plasma diverts current around dense pinch necks before imploding to supply late emission~\citep{ivanov2013,ivanov2015current,ivanov2019diagnostics}.

Together, these comparisons separate three physical effects that single-scalar yield metrics conflate: coherent helical displacement governs bulk compression and energy retention, multimode axial fragmentation softens the $>1$~keV spectrum, and trailing-plasma breadth controls current shunting and radiation release timing, as detailed with the Stage~III coarse histories in Appendix~\ref{app:alresponse}.

\subsection{System behavior}
\label{sec:alroles}

Across $136.3$~h in two execution periods, the investigation recorded $1{,}196$ computational executions: $974$ succeeded, $194$ failed, and $28$ were cancelled, alongside $131$ independent method reviews across four experimental stages. The agents proved effective at sustained numerical engineering and single-pass verification: they diagnosed a silent three-dimensional transport failure, cut three-dimensional runtime by $62.9\%$ through MPI decomposition and radiation-solver tuning, implemented and verified the single-helix and multimode Fortran initializers, cross-checked reconstructed diagnostics against native solver ledgers, and executed fine-mesh timestep-contraction controls while preserving all trajectories. Conversely, when single-scalar tests encountered coupled multi-mechanism physics, four brief human steering interventions supplied the physical taste defined in Sec.~\ref{sec:introduction} to shift the regime from linear seeds to helical kinking, multimode fragmentation, and trailing-plasma current shunting. Within those steered regimes, the agents uncovered the inverse trend between helical Ohmic heating and radiation yield in Stage~II, isolated the trailing-plasma delay in Stage~III, and executed the prospective fine-grid timing test in Stage~IV, after which author analysis reconstructed the full field and load-versus-background energy histories.

Stages~I--III ran on a 16-core host using a DeepSeek Flash worker with GPT-6.1 Sol as director and reviewers, whereas Stage~IV ran 32-rank fine-grid calculations with GPT-6.1 Terra as worker, GPT-6 Astra as director, GPT-6.1 Sol as reviewer, and the Conjecture~1 coprocessor providing live supervisory routing. Table~\ref{tab:alresources} reports the combined token usage across all four stages after deduplicating cumulative conversation receipts by role and thread; Appendix~\ref{app:studyconfig} details the stage configurations.

\begin{table}[tbp]
\centering
\caption{Recorded language-model token usage in millions across all four Z-pinch stages. Cached input is included in total input; the reviewer row includes report assessment.}
\label{tab:alresources}
\begin{ruledtabular}
\begin{tabular}{lrrr}
Role & Input & Cached input & Output \\
\colrule
Worker & 1,042.693 & 1,005.254 & 8.933 \\
Director & 169.151 & 129.747 & 1.365 \\
Reviewer & 119.934 & 102.835 & 0.916 \\
Memory & 0.347 & 0.197 & 0.016 \\
\colrule
Total & 1,332.125 & 1,238.032 & 11.230 \\
\end{tabular}
\end{ruledtabular}
\end{table}

\section{Summary and outlook}
\label{sec:discussion}
\label{sec:extensions}
\label{sec:conclusion}

This work presents \simj{}, an autonomous research framework that centers simulation campaigns on falsification, counterexample search, and prospective hypothesis repair. Separating gated instrument qualification from scientific claim review allows language-model agents to adapt solvers, physical closures, and diagnostics during a study without confounding numerical artifacts with physical mechanisms. In the two-dimensional resistive-MHD validation study, this workflow autonomously isolated and verified a counterexample to a Sweet--Parker reconnection-rate scaling conjecture. In the $136.3$~h three-dimensional aluminum Z-pinch investigation, combining autonomous multi-resolution execution with human steering and analysis showed that stronger coherent helical deformation weakens transverse compression and leaves more magnetic and thermal energy stored despite larger resistive conversion, whereas broadening the trailing-plasma profile shunts axial current early and delays core assembly. In the fine-grid intermediate-width comparison, additional late emission nearly closes the early bolometric deficit by $44$~ns, with little change in cumulative hard X-ray yield.

The recorded study trajectories also provide a testbed for evaluating scientific agents on end-to-end research tasks. Existing benchmarks such as SciCode and ScienceAgentBench assess isolated code generation or static data-analysis pipelines~\citep{tian2024scicode,chen2024scienceagentbench}. Constructing replayable diagnostic benchmarks from recorded simulation histories showed that raw numerical coding does not guarantee completed scientific delivery. Models frequently wrote scripts that passed held-out numerical accuracy checks yet still failed the research contract by misreading solver unit conventions, dropping provenance bindings, or exhausting wall-clock budgets before finalizing required verification artifacts. Evaluating agents against recorded simulation stages with independent re-execution separates three capabilities that static coding tests conflate: numerical correctness, scientific judgment when deciding whether a discrepancy survives refinement, and disciplined delivery under fixed resource limits.

A second direction we explored is selective supervision during multi-day investigations. In long campaigns, invoking frontier director or reviewer models on a fixed schedule is token-intensive, yet unmonitored runs can waste hours in late status-polling loops or stalled solver stages. To test whether a small model can route attention more economically, we trained and integrated a compact $4\text{B}$-parameter decision policy, released separately as Conjecture~1~\citep{conjecture1}, that reads compressed workflow state and chooses whether to keep observing quietly, trigger a read-only diagnostic check, or wake the director or review role while leaving scientific approval to the host gates. Across initial workflow trials, this separation showed that lightweight policies can handle routine wakeups and bounded checks, while reward design must balance timely fault recovery against unnecessary interruptions during quiet simulation windows.

Developing the system also required comparing a prescribed classical harness against a minimal harness. In our initial classical mode, the harness enforced a rigid schedule of mandatory plan reviews, stage transitions, and explicit transport bookkeeping before the worker could advance. Development trials showed that this strong harness frequently trapped adequate evidence inside administrative stages and prevented the agent from completing multi-case simulation matrices before the deadline, whereas an unharnessed agent finished the runs quickly but exited without a verified report or prospective repair check. We therefore retired the classical workflow in favor of minimal mode, in which the worker freely owns its experimental plan and native coding tools while the host enforces only durable provenance, source-bound methods review, and guarded claim closure.

These lessons define our next steps in both plasma physics and autonomous research design. Anchoring the three-dimensional model to $1$~MA Zebra aluminum parameters reproduces key experimental mechanisms, including magnetic-reservoir dominance, helical hot-spot localization, and trailing-plasma current shunting that delays stagnation emission. At the same time, the present calculations have two main physical and modeling limitations. First, the prescribed post-ablation initialization and constant $1$~MA drive omit early wire ablation and generator-circuit coupling. Second, single-fluid Eulerian MHD cannot evolve a true vacuum or capture kinetic reconnection physics. The calculation therefore requires a $10^{-3}~\mathrm{kg/m^3}$ background density floor that accumulates significant late-time energy, while single-fluid resistivity conflates reconnecting sheets with background diffusion. Addressing these limits motivates coupled ablation and circuit models, floor-density and vacuum-resistivity controls, and fluid-to-kinetic handoff into particle-in-cell solvers. On the system side, we will extend the minimal evidence harness to additional simulation-gated physics domains while using recorded campaign trajectories to improve agent scientific taste and workflow supervision. We welcome community contributions of scientific conjectures, solver integrations, and benchmark tasks.
\section*{Code and data availability}
The \simj{} source code, study records, reduced energy and radiation ledgers, and figure-reproduction scripts are available in the archived software release~\citep{simjecture}; raw three-dimensional simulation checkpoint arrays are available from the corresponding author upon reasonable request.

\begin{acknowledgments}
We thank Yuanbo Lu for insightful discussions on Z-pinch energy accounting and numerical vacuum treatment.
\end{acknowledgments}

\appendix

\section{Study configurations and numerical controls}
\label{app:studyconfig}

Table~\ref{tab:studyroles} lists the model configurations and supplied starting material across the two investigations, including the role changes in Z-pinch Stage~IV. Reasoning effort settings apply to each role, and every director or reviewer call runs in an isolated context. The reconnection study uses \simj{} v0.7.0, with full source and runtime identities retained in the study records.

\begin{table*}[tbp]
\centering
\caption{Agent configurations and supplied starting material across the two investigations, with the Z-pinch configuration shown by stage.}
\label{tab:studyroles}
\begin{ruledtabular}
\begin{tabular}{p{0.18\linewidth}p{0.18\linewidth}p{0.25\linewidth}p{0.28\linewidth}}
\raggedright Study & \raggedright Worker & \raggedright Director and reviewers & \raggedright Supplied starting material \tabularnewline
\colrule
\raggedright Reconnection scaling & \raggedright GPT-6.1 Sol, medium & \raggedright GPT-6.1 Sol, high & \raggedright FLASH resistive-MHD application, simulation driver, diagnostics, and validation examples. \tabularnewline
\raggedright Z-pinch, Stages I--III & \raggedright DeepSeek Flash & \raggedright GPT-6.1 Sol, high & \raggedright Experimental load and question, FLASH radiation-MHD runtime, aluminum material tables, and commissioning examples. \tabularnewline
\raggedright Z-pinch, Stage IV & \raggedright GPT-6.1 Terra, high & \raggedright Director: GPT-6 Astra, high; reviewers: GPT-6.1 Sol, high & \raggedright Existing FLASH workspace, Stage~I--III trajectories, checkpoints, diagnostics and physical synthesis, and the Conjecture~1 coprocessor. \tabularnewline
\end{tabular}
\end{ruledtabular}
\end{table*}

\subsection{Reconnection-rate measurement and qualification}
\label{app:islandmethods}

Using the initial-state-subtracted reconnected flux $\psi_{\rm rec}$ normalized by $B_0L_0$ and time $t$ normalized by $\tau_A=L_0/v_{A0}$, the worker measures the reconnection rate across $\psi_{\rm rec}\in[0.01,0.05]$ and evaluates the pairwise scaling exponent for $S_a<S_b$:
\begin{equation}
 \begin{aligned}
 R&=\frac{0.04}{t(\psi_{\rm rec}=0.05)-t(\psi_{\rm rec}=0.01)},\\
 p_{\rm eff}&=\frac{\ln[R(S_b)/R(S_a)]}{\ln(S_b/S_a)}.
 \end{aligned}
\end{equation}
Following the exploratory $128^2$ scan across $S_\eta\in\{250,500,1000,2000,4000\}$ and the qualification controls described in Sec.~\ref{sec:island}, the fresh $384^2$ production runs yield $R(1000)=0.0496886$ and $R(2000)=0.0382421$ for $p_{\rm eff}=-0.3778$ in Fig.~\ref{fig:islandfields}(g). Combining the $384^2\to512^2$ spatial-refinement shift and $512^2$ uncertainty at $S_\eta=2000$ with $256^2$ halved-timestep, crossing-bracket, and orthogonal flux-reconstruction sensitivities produces empirical numerical allowances of $1.31\%$ at $S_\eta=1000$ and $2.39\%$ at $S_\eta=2000$, which propagate to $p_{\rm eff}\in[-0.4314,-0.3247]$. Positivity and net conservation hold across the window.

\subsection{Z-pinch initial conditions and diagnostics}
\label{app:alnumerics}

The FLASH calculations use a second-order Harten--Lax--van Leer discontinuities Riemann solver with three-dimensional corner-transport upwind integration and 30 photon groups spanning $0.1$~eV--$100$~keV on an $18\times18\times20~\mathrm{mm^3}$ Cartesian domain. The fine $128^2\times64$ and coarse $64^2\times32$ meshes have transverse cell sizes of $0.141$ and $0.281$~mm, and the baseline radial widths of the core and trailing Gaussian annuli are $0.2$ and $0.4$~mm.

The single-helix initializer combines a $5$~mm-wavelength centerline displacement with a $60\%$ axial density modulation. The multimode initializer combines wavelengths $5$, $10$, and $2.5$~mm with Euclidean-normalized density weights $0.7, 0.2, 0.1$ and displacement weights $0.6, 0.3, 0.1$, using fixed phases and mixed helicities. Table~\ref{tab:alstudy} lists the four Stage~I screening runs, the six Stage~II--III nonlinear displacement cases, and the Stage~IV fine pair, which ran at $\mathrm{CFL}=0.4$ with a radiation linear-solver relative tolerance of $10^{-10}$ and a $4\times4\times2$ MPI decomposition. Initial tagged masses are $76.3408$ and $76.3412~\mu\mathrm{g}$, respectively. After startup, field snapshots are retained approximately every $0.25$~ns.

\begin{table*}[tbp]
\centering
\caption{Principal three-dimensional Z-pinch trajectories over $0$--$32$~ns, where $D$ is the initial centerline displacement amplitude and $\sigma_t$ is the trailing-annulus width.}
\label{tab:alstudy}
\begin{ruledtabular}
\begin{tabular}{llcccrr}
Case & Seed & $D$ & $\sigma_t$ & Mesh & $E_\gamma$ & $E_{>1\,\rm keV}$ \\
 & & (mm) & (mm) & & (kJ) & (J) \\
\colrule
\code{3D-64-L5} & Density, $5\%$ & 0.0 & 0.4 & $64^2\!\times32$ & 3.834 & 34.8 \\
\code{3D-96-L0} & Density, $0\%$ & 0.0 & 0.4 & $96^2\!\times32$ & 3.783 & 85.1 \\
\code{3D-96-L5} & Density, $5\%$ & 0.0 & 0.4 & $96^2\!\times32$ & 3.786 & 82.7 \\
\code{3D-96-L10} & Density, $10\%$ & 0.0 & 0.4 & $96^2\!\times32$ & 3.781 & 83.8 \\
\colrule
\code{3D-128-S4/4} & Helix & 0.4 & 0.4 & $128^2\!\times64$ & 3.678 & 449.4 \\
\code{3D-128-S8/4} & Helix & 0.8 & 0.4 & $128^2\!\times64$ & 3.482 & 439.7 \\
\code{3D-128-M4/4} & Multimode & 0.4 & 0.4 & $128^2\!\times64$ & 3.718 & 417.6 \\
\code{3D-64-S4/4} & Helix & 0.4 & 0.4 & $64^2\!\times32$ & 3.933 & 384.3 \\
\code{3D-64-M4/4} & Multimode & 0.4 & 0.4 & $64^2\!\times32$ & 3.926 & 366.1 \\
\code{3D-64-M4/10} & Multimode & 0.4 & 1.0 & $64^2\!\times32$ & 3.768 & 333.0 \\
\colrule
\multicolumn{7}{l}{Stage~IV, $\mathrm{CFL}=0.4$} \\
\code{3D-128-M4/4} & Multimode & 0.4 & 0.4 & $128^2\!\times64$ & 3.694 & 411.4 \\
\code{3D-128-M4/7} & Multimode & 0.4 & 0.7 & $128^2\!\times64$ & 3.667 & 416.7 \\
\end{tabular}
\end{ruledtabular}
\end{table*}

The centroid-relative RMS radius is
\begin{equation}
 \begin{aligned}
  a_\perp&=\left\langle\left|\mathbf r_\perp-\bar{\mathbf r}_\perp(z)\right|^2\right\rangle_w^{1/2},\\
  w&=\max(\rho-10^{-3}~\mathrm{kg/m^3},0),
 \end{aligned}
 \label{eq:alwidth}
\end{equation}
where $\bar{\mathbf r}_\perp(z)$ is the weighted centroid of each axial slice. Both fine single-helix runs begin at $a_\perp=1.935$~mm. At $32$~ns, the maximum slice-centroid offsets are $2.16$ and $1.28$~mm for \code{3D-128-S8/4} and \code{3D-128-S4/4}, with mid-plane density maxima of $10.8$ and $18.3~\mathrm{kg/m^3}$, peak Ohmic power densities of $0.87$ and $2.19~\mathrm{GW/mm^3}$, and peak stored radiation energy densities of $2.32$ and $6.03~\mathrm{mJ/mm^3}$. Across the Stage~II--III nonlinear scan, the primary power maximum occurs at $4.65$--$5.54$~ns. In the Stage~III \code{3D-128-M4/4} run, mid-plane azimuthal-mean magnetic partitioning between $3$ and $9$~ns shows the enclosed-current fraction in dense plasma with $\rho>0.10\rho_{\max}$ dropping from $44\%$ to $8\%$ as the $90\%$ enclosed-current radius $r_{90}$ reaches $5.6$~mm. The Stage~IV longitudinal cut at $24.03$~ns in Fig.~\ref{fig:alstructure} reaches $\rho_{\max}=33.6~\mathrm{kg/m^3}$, $|\mathbf B|_{\max}=192$~T, and peak Ohmic power density $q_\Omega=3.98~\mathrm{GW/mm^3}$. In \code{3D-64-M4/4}, late-time mass-weighted azimuthal harmonic amplitudes satisfy $A_3=1.93\,A_1$, and the coarse trailing-width pair has $32$~ns RMS centroid offsets of $0.43$ and $0.49$~mm for the broader and narrower profiles.

\subsection{Energy accounting and numerical sensitivity}
\label{app:alenergychecks}

The energy inventories separate kinetic, magnetic, electron, ion, and stored radiation components, subtracting stored radiation from FLASH's combined internal-energy variable to avoid double counting. Table~\ref{tab:alpartition} records the common $0$--$32$~ns comparison, with endpoint inventories interpolated between native end-of-step records and escaping photon energy integrated over exact timestep overlaps, where photon groups 21--30 span $1$--$100$~keV. Across the shared $3.149$--$25.029$~ns stagnation interval of the Stage~III \code{3D-128-M4/4} run, the independently recorded native hydro and diffusion increments sum to $3.245650$~kJ, matching radiation escape plus the net increase of gas and stored radiation energy to within $2\times10^{-5}$~J. The whole-domain cumulative balance is $\mathcal B(t)=E_{\rm stored}(t)-E_{\rm stored}(0)+E_{\gamma,\rm escape}(0,t)$, combining net boundary exchange with numerical contributions. Appendix~\ref{app:vacuumaccounting} relates these inventories to the load-and-field partition in Fig.~\ref{fig:alenergy}.

The Stage~IV reference in Figs.~\ref{fig:alstructure} and~\ref{fig:alenergy} uses this same $3.149$--$25.029$~ns evaluation window; its compression minimum occurs at $25.27$~ns. Over the fixed window, reconstructed resistive conversion is $329.78$~J versus a native diffusion-stage gas-energy increment of $330.16$~J, a $0.12\%$ difference; retaining every second field snapshot changes the reconstruction by $0.0074\%$. The higher output cadence places the kinetic-reference sample at $9.776$~ns, while the Stage~III resolution comparisons keep their original reference samples. In the Stage~IV per-rank hydrodynamic logs, the recorded hydrodynamic energy adjustment contributes $+61.55$~J over the stagnation window and $+13.01$~J over $0$--$32$~ns in \code{3D-128-M4/4 (IV)}, compared with $+58.88$ and $+10.64$~J in \code{3D-128-M4/7}. Summing the conservative update with this adjustment reproduces the native hydrodynamic energy change with a summed absolute discrepancy below $7\times10^{-5}$~J over each $0$--$44$~ns history.

\begin{table*}[tbp]
\centering
\caption{Whole-domain initial stored energy $E_0$, $32$~ns stored components $E_B$, $K$, $U_g=U_e+U_i$, and $U_\gamma$, cumulative escaping radiation $E_\gamma$, and net boundary-and-residual balance $\mathcal B=E_B+K+U_g+U_\gamma+E_\gamma-E_0$, in kJ.}
\label{tab:alpartition}
\begin{ruledtabular}
\begin{tabular}{lrrrrrrr}
Case & $E_0$ & $E_B$ & $K$ & $U_g$ & $U_\gamma$ & $E_\gamma$ & $\mathcal B$ \\
\colrule
\code{3D-128-S4/4} & 4.298 & 0.791 & 0.607 & 0.593 & 0.002 & 3.678 & 1.374 \\
\code{3D-128-S8/4} & 4.293 & 0.887 & 0.581 & 0.706 & 0.002 & 3.482 & 1.364 \\
\code{3D-128-M4/4 (III)} & 4.298 & 0.786 & 0.555 & 0.593 & 0.002 & 3.718 & 1.357 \\
\code{3D-64-S4/4} & 4.290 & 0.665 & 0.550 & 0.420 & 0.003 & 3.933 & 1.281 \\
\code{3D-64-M4/4} & 4.290 & 0.654 & 0.543 & 0.419 & 0.003 & 3.926 & 1.256 \\
\code{3D-64-M4/10} & 4.286 & 0.609 & 0.608 & 0.393 & 0.004 & 3.768 & 1.096 \\
\colrule
\code{3D-128-M4/4 (IV)} & 4.298 & 0.766 & 0.577 & 0.653 & 0.002 & 3.694 & 1.394 \\
\end{tabular}
\end{ruledtabular}
\end{table*}

In the Stage~III resolution audit, reconstructed Ohmic conversion agrees with the native diffusion increment to $2.1\times10^{-4}$ relative on the fine mesh, with $86.9\%$ inside $r<2$~mm and $87.0\%$ of the summed absolute fine--coarse mesh difference inside $1$~mm. While refining from $64^2\!\times32$ to $128^2\!\times64$ shifts absolute yields toward harder emission by resolving hotter pinched segments, matched comparisons at fixed resolution preserve the sign and ordering of both the single-to-multimode spectral softening and the trailing-width delay across both grids (Table~\ref{tab:alstudy}). In a separate window scan on the extended $46.04$~ns fine trajectory that uses whole-domain kinetic energy immediately before the hydro update at each window's leading edge, power thresholds $f\in\{0.75,0.50,0.25,0.20,0.10\}$ yield excesses of $31.4$~J, $318.8$~J, $3.033$~kJ, $3.283$~kJ, and $3.684$~kJ. Over $0$--$32$~ns, the native hydro-stage net gas-energy increments for \code{3D-128-S4/4} and \code{3D-128-S8/4} are $3.719$ and $3.518$~kJ, and all snapshot magnetic and kinetic inventories agree with the native histories within $2\times10^{-6}$ relative tolerance. In the Stage~IV fine-pair controls, continuing from $32$ to $44$~ns at $\mathrm{CFL}=0.2$ gives $27.87$~J more radiation in the wider case versus $30.49$~J at $\mathrm{CFL}=0.4$, while local checkpoint continuations from $29.52$ to $32.5$~ns at $\mathrm{CFL}=0.2$, $0.1$, and $0.05$ contract the $32$~ns gas-plus-radiation energy shift from $15.22$ to $7.79$~J in the narrow case and from $13.52$ to $6.98$~J in the wider case, measuring the local timestep sensitivity of the evolved gas-plus-radiation inventories. Across both full Stage~IV histories, the stepwise decrease of gas-plus-radiation energy during each radiation update matches the recorded escaping radiation with a maximum residual of $2.4~\mu\mathrm{J}$ and a summed absolute residual below $7~\mu\mathrm{J}$.

\subsection{Retained current and radiation histories}
\label{app:alresponse}

The author analysis uses $565$ retained three-dimensional field snapshots from seven nonlinear trajectories: $219$ in Stages~II--III and $346$ in Stage~IV, together with their complete radiation ledgers and the Stage~I screening ledgers in Table~\ref{tab:alstudy}. Every native radiation step in the original $0$--$32$~ns extracts is reproduced by the longer records without gaps or overlaps, and each escaped group energy equals its power times the native timestep. The common continuation comparison stops at $44$~ns, before the shorter trailing-profile trajectory ends. Spectrograms in Fig.~\ref{fig:alradiation} retain the original 30 photon groups without smoothing, where color represents group power divided by logarithmic energy width, gray marks spectral power below $0.01$~GW per logarithmic interval, and integrated spectra use exact timestep overlaps with group centers connected directly. The $>1$~keV sum includes both continuum and line emission, and each nonlinear case has a single interior local power maximum between $10$ and $32$~ns with prominence above $10$~GW. Table~\ref{tab:alintervalradiation} lists the interval yields and hard fractions for the Stage~III and~IV trailing-width pairs.

\begin{table}[tbp]
\centering
\caption{Interval radiation yield $E_\gamma$ and hard-band fraction $f_{\rm hard}=E_{>1\,\rm keV}/E_\gamma$ for the Stage~III and Stage~IV trailing-width pairs.}
\label{tab:alintervalradiation}
\begin{ruledtabular}
\begin{tabular}{lrrrr}
Interval & \multicolumn{2}{c}{$E_\gamma$ (kJ)} & \multicolumn{2}{c}{$f_{\rm hard}$ (\%)} \\
(ns) & Narrow & Broad & Narrow & Broad \\
\colrule
\multicolumn{5}{l}{Stage~III: $0.4/1.0$~mm, $64^2\!\times32$} \\
0--8 & 1.012 & 0.985 & 17.81 & 17.71 \\
8--16 & 0.983 & 0.946 & 7.50 & 3.88 \\
16--24 & 1.203 & 0.957 & 6.31 & 6.47 \\
24--32 & 0.728 & 0.880 & 4.97 & 6.79 \\
32--44 & 0.693 & 0.789 & 4.14 & 4.73 \\
\colrule
\multicolumn{5}{l}{Stage~IV: $0.4/0.7$~mm, $128^2\!\times64$} \\
0--8 & 1.055 & 0.992 & 18.45 & 18.69 \\
8--16 & 0.976 & 0.961 & 10.10 & 7.38 \\
16--24 & 1.080 & 1.049 & 7.79 & 9.98 \\
24--32 & 0.583 & 0.664 & 5.85 & 8.37 \\
32--44 & 0.455 & 0.485 & 6.46 & 6.20 \\
\end{tabular}
\end{ruledtabular}
\end{table}

Axial current in Fig.~\ref{fig:alcurrentresponse}(a) is reconstructed from $J_z=(\partial_xB_y-\partial_yB_x)/\mu_0$ using native-grid central differences and the verified FLASH magnetic-field normalization. At each axial slice, the numerator of the inner-current fraction integrates signed $J_z$ within $2$~mm of that slice's density-weighted centroid; the denominator integrates within $8$~mm of the original axis. Currents are averaged over axial slices before taking their ratio. For the Stage~III pair, the narrow and broad profiles enclose mean inner currents of $364$ and $304$~kA over $12$--$24$~ns, giving signed fractions of $0.536$ and $0.402$; counting only forward current gives $0.544$ and $0.421$. Density-defined cores at $5\%$, $10\%$, and $20\%$ of each slice's peak density give the same ordering.

The hot-load diagnostic in Fig.~\ref{fig:alcurrentresponse}(b) integrates tracer-weighted mass where $T_e>300$~eV and $\rho\ge0.01~\mathrm{kg/m^3}$. For the Stage~III pair, following the initial transient, the first retained snapshot after $10$~ns with more than $0.1\%$ of the load above that threshold occurs at $13.11$~ns in the narrow case and $18.02$~ns in the broad case, with preceding samples at $12.01$ and $17.01$~ns, while the interior stagnation power maximum shifts from $18.31$ to $22.89$~ns, a delay of $4.58$~ns. The Stage~III broad profile has a $158.6$~J bolometric deficit at $32$~ns and recovers $60\%$ of that deficit by $44$~ns, leaving a $1.37\%$ total-yield difference and a $6.21\%$ lower cumulative $>1$~keV yield.

For the Stage~IV fine-grid pair with $\sigma_t=0.4$ and $0.7$~mm across $173$ unique field times per case, the first post-$10$~ns crossing at which $0.1\%$ of tagged mass exceeds $300$~eV lies in $[12.53,12.77]$ and $[14.02,14.28]$~ns, giving a linearly interpolated delay of $1.64$~ns. The $500$~eV crossing is delayed by $1.73$~ns, whereas the $1$~keV crossing occurs $1.26$~ns earlier in the wider case. Minimum centroid-relative radii of $0.514$ and $0.492$~mm occur at $25.27$ and $27.03$~ns. Applying the fixed peak rule to $0.1$~ns radiation bins with $0.5$~ns centered smoothing places the stagnation power maxima at $17.55$ and $19.35$~ns, a $1.80$~ns delay that is unchanged across $0.2$~ns bins and $0.3$ or $0.7$~ns smoothing, with unsmoothed $0$--$44$~ns yields of $4.1483$ and $4.1517$~kJ and $>1$~keV yields of $440.8$ and $446.7$~J.

\section{Numerical vacuum treatment and energy accounting}
\label{app:vacuumaccounting}

The Eulerian MHD setup represents the exterior vacuum region with a low-density floor material, \code{vacu}, of mass fraction $X_v$. The initial density combines the core shell and trailing annulus, which share the load tracer and total $76.34~\mu\mathrm{g}$, with a uniform $10^{-3}~\mathrm{kg/m^3}$ background contributing $6.48~\mu\mathrm{g}$ across the $18\times18\times20~\mathrm{mm^3}$ domain. In the single-fluid update, all species share the fluid velocity and electron and ion temperatures, while the background follows an ideal-gas equation of state with $\gamma_e=\gamma_i=5/3$, molar mass $A_v=26.9815385~\mathrm{g/mol}$, fixed charge $Z_v=13$, zero opacity and emissivity, and no vacuum-resistivity override.

The background material energies follow from the recorded mass fraction and its equation of state,
\begin{equation}
 \begin{aligned}
 K_v&=\int_V \tfrac12\rho X_v|\mathbf v|^2\,\dd V,\\
 U_v&=\int_V\rho X_v\frac{R}{A_v}
       \left(\frac{T_i}{\gamma_i-1}+\frac{Z_vT_e}{\gamma_e-1}\right)\dd V,
 \end{aligned}
 \label{eq:vacuumenergy}
\end{equation}
where $R$ is the universal gas constant. Subtracting the background components separately for electrons and ions gives the additive load inventories $K_{\rm L}=K-K_v$ and $U_{g,\rm L}=U_g-U_v$ across all $173$ retained field snapshots per Stage~IV trajectory. Reconstructed background thermal energies agree with native specific energies in nearly pure background cells to within $3.2\times10^{-6}$ relative, and interpolating the background integrals to the native ledger cadence changes the nominal $32$~ns kinetic, electron, and ion background inventories by less than $0.04$~J when subsampled at every second snapshot.

\begin{figure*}[tbp]
\centering
\includegraphics[width=\linewidth]{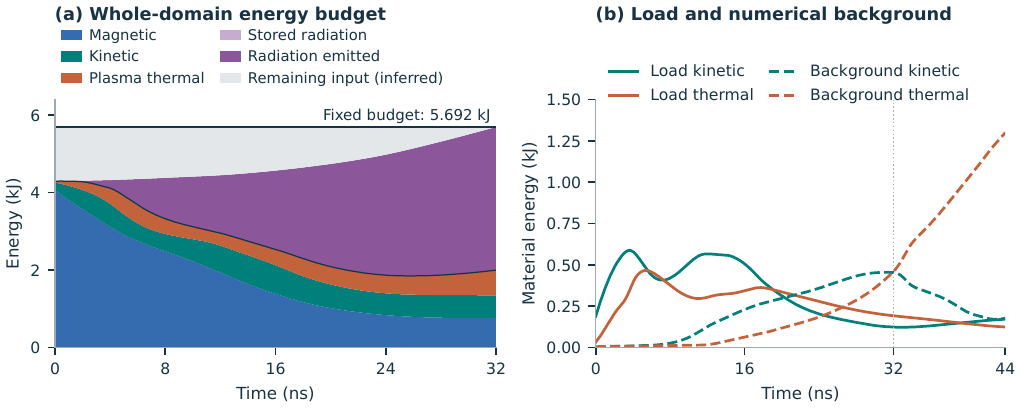}
\caption{Energy accounting for \code{3D-128-M4/4 (IV)}. (a) Whole-domain budget over $0$--$32$~ns, including the numerical background. (b) Separate kinetic and thermal energies of the aluminum load (solid) and numerical background (dashed) through $44$~ns. Thermal energy includes electrons and ions.}
\label{fig:alvacuum}
\end{figure*}

Figure~\ref{fig:alvacuum} shows the whole-domain budget and the separate material histories. In the nominal Stage~IV run, background kinetic energy reaches $0.454$~kJ and background thermal energy reaches $0.461$~kJ at $32$~ns, while the load retains $0.124$ and $0.192$~kJ, respectively. By $44$~ns, the background material holds $1.476$~kJ in total. Subtracting $E_v=K_v+U_v$ defines the retained inventory and its net exchange,
\begin{equation}
 \begin{aligned}
 E_{\rm L}&=E_B+K_{\rm L}+U_{g,\rm L}+U_\gamma,\\
 \mathcal B_{\rm L}(t)
  &=E_{\rm L}(t)-E_{\rm L}(0)+E_{\gamma,\rm escape}(0,t)\\
  &=\mathcal B(t)-[E_v(t)-E_v(0)].
 \end{aligned}
 \label{eq:loadbalance}
\end{equation}
Magnetic and radiation-field energies retain their full spatial integrals, including the nominal vacuum region. The gray band in Fig.~\ref{fig:alenergy} is therefore the remaining net exchange with the retained load and fields; energy accumulated in background material enters the excluded part of that exchange. At $32$~ns, this gives $\mathcal B_{\rm L}=0.485$~kJ and the fixed budget $E_{\rm L}(0)+\mathcal B_{\rm L}(32~\mathrm{ns})=4.777$~kJ. The whole-domain inventories in Table~\ref{tab:alpartition}, the operator audits in Appendix~\ref{app:alenergychecks}, and the volume-integrated $Q_\Omega$ retain their original definitions. Both Stage~IV trailing-width runs share this identical background floor, while quantifying sensitivity to the floor density and vacuum resistivity requires separate numerical controls.

\bibliographystyle{apsrev4-2}
\makeatletter
\immediate\write\@auxout{\string\citation{apsrev42Control}}
\makeatother
\let\simjBibitemNoStop\bibitemNoStop
\renewcommand{\bibitemNoStop}{\unskip\simjBibitemNoStop}
\makeatletter
\preprintsty@sw{}{%
 \renewcommand{\bibsection}{%
  \par
  \begingroup
   \baselineskip26\p@
   \bib@device{\hsize}{72\p@}%
  \endgroup
  \nobreak\@nobreaktrue
  \addvspace{19\p@}%
 }%
}
\makeatother
\bibliography{references}

\end{document}